\documentclass[sigconf]{acmart}

\AtBeginDocument{%
  \providecommand\BibTeX{{%
    \normalfont B\kern-0.5em{\scshape i\kern-0.25em b}\kern-0.8em\TeX}}}

\copyrightyear{2021}
\acmYear{2021}
\setcopyright{rightsretained}
\acmConference[MRC'21]{The 1st International Workshop on Machine Reasoning}{March 12, 2021}{Jerusalem, Israel}
\acmBooktitle{Proceedings of the 1st International Workshop on Machine Reasoning (MRC'21), March 12, 2021, Jerusalem, Israel}
\acmDOI{}
\acmISBN{}

\usepackage{tikz}
\usepackage{subcaption}

\begin{document}

\title{Wider Vision: Enriching Convolutional Neural Networks via Alignment to External Knowledge Bases}


\author {
    Xuehao Liu
}
\affiliation{%
    \institution{Technological University Dublin}
    \city{Dublin}
    \country{Ireland}
}
\email{xuehao.liu@tudublin.ie}

\author {
    Sarah Jane Delany 
}
\affiliation{%
    \institution{Technological University Dublin}
    \city{Dublin}
    \country{Ireland}
}
\email{sarahjane.delany@tudublin.ie}

\author {
    Susan McKeever 
}
\affiliation{%
    \institution{Technological University Dublin}
    \city{Dublin}
    \country{Ireland}
}
\email{susan.mckeever@tudublin.ie}

\renewcommand{\shortauthors}{Liu, et al.}

\begin{abstract}
    
    Deep learning models suffer from opaqueness. For Convolutional Neural Networks (CNNs), current research strategies for explaining models focus on the target classes within the associated training dataset. As a result, the understanding of hidden feature map activations is limited by the discriminative knowledge gleaned during training. The aim of our work is to explain and expand CNNs models via the mirroring or alignment of the network to an external knowledge base. This will allow us to give a semantic context or label for each visual feature.  Using the resultant aligned embedding space, we can match CNN feature activations to nodes in our external knowledge base. This supports knowledge-based interpretation of the features associated with model decisions. 
    

        To demonstrate our approach, we build two separate graphs from:  (1) ConceptNet knowledge base and (2) a public CNN. We use an entity alignment method to align the feature nodes in the CNN with the nodes in the ConceptNet based knowledge graph. We then measure the proximity of CNN graph nodes to semantically meaningful knowledge base nodes.  Our results show that in the aligned embedding space, nodes from the knowledge graph are close to the CNN feature nodes that have similar meanings, indicating that nodes from an external knowledge base can act as explanatory semantic references for features in the model. We analyse a variety of graph building methods in order to improve the results from our embedding space. We further demonstrate that by using hierarchical relationships from our external knowledge base, we can locate new unseen classes outside the CNN training set in our embeddings space based on visual feature activations. This suggests that we can adapt our approach to identify unseen classes based on CNN feature activations.   
    Our demonstrated approach of aligning a CNN with an external knowledge base paves the way to reason about and beyond the trained model, with future adaptations to explainable models and zero-shot learning. 
    
\end{abstract}





\maketitle

\section{Introduction}

  
  Deep Learning models are considered to be black boxes, with a lack of visibility of layers and decisions paths \cite{castelvecchi2016can}. With the increasing complexity of models, research has focused on providing explanations to explain the relationship between model inputs and outputs \cite{arrieta2020explainable,gilpin2018explaining}. Explanations can be used for several purposes. One purpose is to underpin the fine-tuning and achievement of higher accuracies \cite{kim2016examples,koh2017understanding}. Another is to help researchers identify why an image input has been incorrectly interpreted \cite{selvaraju2017grad}. Other work in the field has shown that different layers of the CNNs interpret different levels of visual features within the image  \cite{simonyan2013deep,zeiler2014visualizing,springenberg2014striving,gatys2016image}.
  Current explanations of CNNs that interpret predictions are limited to interpreting the relationship between inputs, feature activation and outputs generated from the training dataset. There is no additional insight or knowledge available outside the knowledge embedded in the training dataset.  \citet{selvaraju2017grad} highlighted the areas of features that make most contribution to the output label. However, this work was unable to show the reason why these areas are combined together. \citet{bau2017network} involved more concepts to explain features that CNNs are detecting, such as textures, colours, and scenes. They showed that specific hidden units in the network are detecting particular semantically meaningful concepts. However, a richer explanation for features  from a wider pool of knowledge beyond the training set is still missing. CNN explanations with no external knowledge are able to indicate which individual features are activated for outputs, but will be unaware of the semantic concepts behind these features, or indeed, how they relate to other outputs. A richer knowledge driven explanation (akin to a human who has real world knowledge) could show, for example, that ear, paw and tail features were appearing in  a particular dog image; that dogs generally have such features, and indeed further reason that these features may appear for other previous unseen animal types to help identify them.  \citet{mitsuhara2019embedding} use attention maps and manual image labelling in the direction of such fine-tuned  explanations. However, manual labelling is not practical for large image datasets. 
  
  Inspired by the work of  \citet{lecue2019role}, our approach is based upon aligning a CNN to an external knowledge base. We hypothesise that by aligning a trained CNN model to an external knowledge base, we can discover semantically meaningful labels for activated CNN  features.  We also hypothesise that we can use the alignment process to identify an unseen output (i.e. a class not included in the training data) based on CNN feature activations.
  
  We represent the CNN and the external knowledge base as two separate graph structures. The premise is that outputs in the CNN (e.g. dog) and their associated activated features align with concepts in the knowledge graph (i.e. dog) and its associated connected nodes (e.g. tail, paws, ears). We then align these two graphs, to create a shared embedding space. We can then locate labels for CNN feature activations in the external knowledge base, via distance measurement in the embedding space. This labelling supports an explanation of a CNN decision. It also paves the way for detecting unseen classes in our CNN, via identification from activated features mapped to the external knowledge base. 
  
 We demonstrate and test our approach using the external knowledge base, ConceptNet \cite{speer2017conceptnet}. For the CNN as a graph, we use VGG-16 \cite{simonyan2014very,zeiler2014visualizing} where the feature layers represent visual features.  
 We align features of the CNN and entities in the knowledge graph using an entity alignment method, using the GCN-Align method \cite{wang2018cross}. 
  
  Our evaluation focuses on two parts (1) whether we have correctly aligned the CNN and knowledge graph and (2) the effectiveness of our embedding space for labelling CNN features. The remainder of the paper is structured into Related Work, our Approach, explained in detail, and our Experimental evaluation and results. In our conclusion, we cover possible future work directions.

\section{Related work}

The closest domain to our work is Explainable AI, as we are aiming to discover the meanings of CNN features. For CNNs, explainable AI techniques started with intuiting what features are being detected in an image.  One way of visualizing features is to trace the gradients back to the input image \cite{zeiler2014visualizing,springenberg2014striving}. The resultant highlighting contour on the input image shows the pattern that the filter is detecting. It also shows that later layers closer to the output will detect higher object level patterns, while earlier network layers detect features that are more basic, such as textures. Tracing back the gradients not only provides a visualization of features, but also highlights the part of an image that leads to the CNN prediction, such as Deconvolution \cite{zeiler2014visualizing,simonyan2013deep}, Layer-Wise Relevance Propagation (LRP) \cite{bach2015pixel}, DeepSHAP\cite{lundberg2017unified}, and Grad-CAM \cite{selvaraju2017grad}. Another explanation approach is to generate an image from noise, and the image maximizes the prediction score \cite{mordvintsev2015inceptionism,simonyan2013deep}. Although such images may not resemble any meaningful object, the patterns found by the CNNs are informative. Generally, while the visualisation of detected features can highlight what and where CNNs have focused in an image,  the meaning of the feature is still unknown until it is interpreted further.  \citet{bau2017network} advanced this interpretation via labelled bounding boxes around concepts within the dataset images. By comparing the activated area of a feature with the ground truth areas, activated features can be identified as specific concepts within an image. 

These explanation approaches can only find explanations based on knowledge gleaned from the training dataset. To find an explanation in a broader knowledge space, we combine the CNN with a knowledge graph in order to assign and enhance feature meanings. We note that this approach of bringing knowledge graphs into machine learning systems in order to expand knowledge beyond the training set is gaining traction in the machine learning research domain - and is applied to both image classification and zero-shot learning. For image classification, \citet{marino2016more} used labels of the COCO dataset \cite{lin2014microsoft} to build a knowledge graph. Both images and concepts of the COCO dataset are from the lexical database, WordNet \cite{miller1995wordnet}. They applied a Graph Search Neural Network(GSNN) to do reasoning through the graph, and thus improved mean average precision. In zero-shot learning, \citet{lee2018multi} also built a knowledge graph using WordNet. Using the features of ResNet \cite{he2016deep} as input, they used a GSNN to infer unknown classes. \citet{wang2018zero} used WordNet,  Never-Ever-Language-Learning (NELL) \cite{carlson2010toward} and Never-Ever-Image-Learning (NEIL) \cite{chen2013neil} to build a knowledge graph, where this graph takes Glove \citep{pennington2014glove} word embeddings as input. They applied a Graph Convolutional Network(GCN) on the graph to generate the weight of the last layer of ResNet, so that this updated ResNet model can then recognise the nodes in the knowledge graph. In Zero-shot Action Recognition, \citet{gao2019know} used ConceptNet \cite{speer2017conceptnet} to build a knowledge graph. In their approach, both Word2vec \cite{mikolov2013distributed} and GoogleNet \cite{szegedy2015going} features are used in parallel as input. Two-Stream Graph Convolutional Network (TS-GCN) is applied on the graph to recognize actions. 
\\The above research indicates that using knowledge graphs as external assistants can benefit and expand the knowledge of current deep learning systems. In our case, where we are linking a CNN to an external knowledge base for feature labelling and network insights, We require \textit{ entity alignment} methods to correctly map our CNN to the knowledge base. Entity alignment is about aligning nodes in two separate graphs that have the same semantic meaning. The most intuitive way of matching two entities is to compare the traditional language features \cite{akbari2009improved}, such as n-grams and TF-IDF\cite{sun2015comparative}. Recently, with the help of node embedding \cite{bordes2013translating}, entity alignment has been done by minimizing the distance between embeddings of labelled nodes with same meanings from two separate graphs \cite{hao2016joint}. In addition, embeddings of the \textit{attributes} for each node (from the graph) can be also used for minimizing distance between two nodes \cite{trisedya2019entity,sun2017cross}.  \citet{sun2018bootstrapping} introduced a bootstrapping strategy to consider nodes that have a distance small enough as the labelled nodes when more training steps are made. The bootstrapping strategy provided a better alignment accuracy on Hit@1 score. All of these methods can align nodes in two graphs. We apply GCN-Align\cite{wang2018cross} as our Entity Alignment method as it could do alignment only using structure information such as neighbours and relationships.

\section{Approach}

\usetikzlibrary{arrows,shapes,snakes,automata,backgrounds,petri}
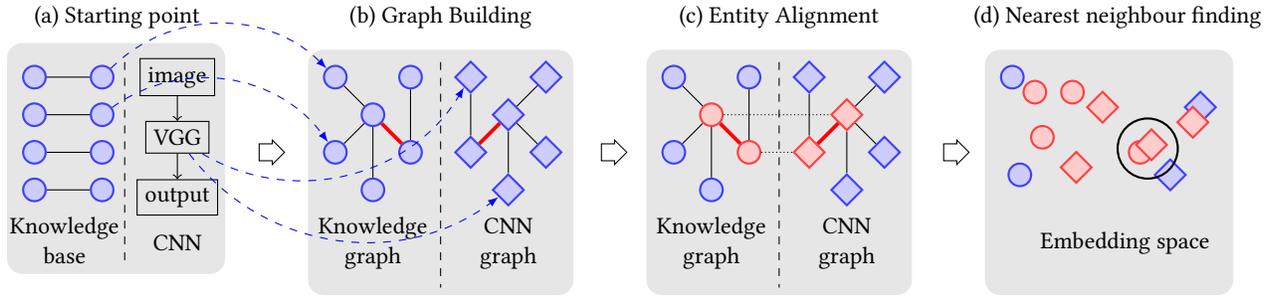
\begin{figure*}[h]\centering
\begin{tikzpicture}[node distance=0.9cm,bend angle=45,auto,edge_style/.style={draw=black, ultra thick}]
  
  \tikzstyle{place}=[circle,thick,draw=blue!75,fill=blue!20,minimum size=3mm]
  \tikzstyle{red place}=[place,draw=red!75,fill=red!20]
  \tikzstyle{cnn}=[diamond,thick,draw=blue!75,fill=blue!20,minimum size=3mm]
  \tikzstyle{red cnn}=[cnn,draw=red!75,fill=red!20]
  \tikzstyle{highlight}=[circle,thick,draw=black,minimum size=8mm]
  \tikzstyle{transition}=[rectangle,thick,draw=black!75,
  			  fill=black!20,minimum size=4mm]

 \begin{scope}
    \node [place] at (0, 0) (a1)                                {};
    \node [place] (a2) [right of=a1]                     {};
    \node [place] at (0, -0.5)(b1) []                    {};
    \node [place] (b2) [right of=b1]                     {};
    \node [place] at (0, -1.0)(c1)                       {};
    \node [place] (c2) [right of=c1]                     {};
    \node [place] at (0, -1.5)(d1)                      {};
    \node [place] (d2) [right of=d1]                     {};
    \draw (a1) -- (a2);
    \draw (b1) -- (b2);
    \draw (c1) -- (c2);
    \draw (d1) -- (d2);
    \draw [dashed] (1.2,0.2) -- (1.2,-2.5);
    \node[rectangle,draw] (image1) at (1.9, 0) {image};
    \node[rectangle,draw] (CNN1) at (1.9, -0.8) {VGG};
    \node[rectangle,draw] (output1) at (1.9, -1.6) {output};
    \draw [->] (image1) -- (CNN1);
    \draw [->] (CNN1) -- (output1);
    \node (title1) at (1.1, 0.8) {(a) Starting point};
    \node (CNt1) at (0.4, -2.2) {\begin{tabular}{cc}Knowledge \\base\end{tabular}};
    \node (CNNt1) at (1.9, -2.2) {CNN};
  \end{scope}

  \begin{scope}[xshift=4cm]
    \node [place] at (0, 0) (e1)                                {};
    \node [place] at (0.5, -0.5)(e2) []                    {};
    \node [place] at (0, -1.0)(e3)                       {};
    \node [place] at (1, -1.0)(e4)                      {};
    \node [place] at (0.5, -1.5)(e5)                      {};
    \node [place] at (1.0, 0)(e6) []                    {};
    \draw (e1) -- (e2);
    \draw (e2) -- (e3);
    \draw [red, line width=0.5mm] (e2) -- (e4);
    \draw (e2) -- (e5);
    \draw (e4) -- (e6);
    \draw [dashed] (1.4,0.2) -- (1.4,-2.8);
    \node [cnn] at (2.8, 0) (e7)                                {};
    \node [cnn] at (2.3, -0.5)(e8) []                    {};
    \node [cnn] at (2.8, -1.0)(e9)                       {};
    \node [cnn] at (1.8, -1.0)(e10)                      {};
    \node [cnn] at (2.3, -1.5)(e11)                      {};
    \node [cnn] at (1.8, 0)(e12) []                    {};
    \draw (e7) -- (e8);
    \draw (e8) -- (e9);
    \draw [red, line width=0.5mm] (e8) -- (e10);
    \draw (e8) -- (e11);
    \draw (e10) -- (e12);
    \node (title2) at (1.4, 0.8) {(b) Graph Building};
    \node (CNt2) at (0.5, -2.2) {\begin{tabular}{cc}Knowledge \\graph\end{tabular}};
    \node (CNNt2) at (2.3, -2.2) {\begin{tabular}{cc}CNN \\graph\end{tabular}};
  \end{scope}

\begin{scope}[xshift=8.5cm]

    \node [place] at (0, 0) (f1)                                {};
    \node [red place] at (0.5, -0.5)(f2) []                    {};
    \node [place] at (0, -1.0)(f3)                       {};
    \node [red place] at (1, -1.0)(f4)                      {};
    \node [place] at (0.5, -1.5)(f5)                      {};
    \node [place] at (1.0, 0)(f6) []                    {};
    \draw (f1) -- (f2);
    \draw (f2) -- (f3);
    \draw [red, line width=0.5mm](f2) -- (f4);
    \draw (f2) -- (f5);
    \draw (f4) -- (f6);
    \draw [dashed] (1.4,0.2) -- (1.4,-2.8);
    \node [cnn] at (2.8, 0) (f7)                                {};
    \node [red cnn] at (2.3, -0.5)(f8) []                    {};
    \node [cnn] at (2.8, -1.0)(f9)                       {};
    \node [red cnn] at (1.8, -1.0)(f10)                      {};
    \node [cnn] at (2.3, -1.5)(f11)                      {};
    \node [cnn] at (1.8, 0)(f12) []                    {};
    \draw (f7) -- (f8);
    \draw (f8) -- (f9);
    \draw [red, line width=0.5mm](f8) -- (f10);
    \draw (f8) -- (f11);
    \draw (f10) -- (f12);
    \draw [densely dotted] (f2) -- (f8);
    \draw [densely dotted] (f4) -- (f10);
    \node (title3) at (1.4, 0.8) {(c) Entity Alignment};
    \node (CNt3) at (0.5, -2.2) {\begin{tabular}{cc}Knowledge \\graph\end{tabular}};
    \node (CNNt3) at (2.3, -2.2) {\begin{tabular}{cc}CNN \\graph\end{tabular}};
  \end{scope}
  
  \begin{scope}[xshift=13cm]
    \node [place] at (0, 0) (g1)                                {};
    \node [red place] at (0.3, -0.2)(g2) []                    {};
    \node [place] at (0.1, -1.3)(g3)                       {};
    \node [red place] at (1.7, -1.0)(g4)                      {};
    \node [red place] at (0.8, -0.2)(g5)                      {};
    \node [red place] at (0.4, -0.8)(g6) []                    {};
    \node [cnn] at (2.5, -0.4) (g7)                                {};
    \node [red cnn] at (2.4, -0.6)(g8) []                    {};
    \node [cnn] at (2.1, -1.3)(g9)                       {};
    \node [red cnn] at (0.85, -1.2)(g10)                      {};
    \node [red cnn] at (1.2, -0.4)(g11)                      {};
    \node [red cnn] at (1.85, -0.9)(g12) []                    {};
    \node [highlight] at (1.8, -0.95)(g13) []                    {};

    \node (title3) at (1.4, 0.8) {(d) Nearest neighbour finding};
    \node (CNNt4) at (1.5, -2.2) {Embedding space};
  \end{scope}
  \draw [dashed, -latex, blue] (a2) to [bend left] (e1);
  \draw [dashed, -latex, blue] (b2) to [bend left] (e3);
  \draw [dashed, -latex, blue] (CNN1) to [bend right] (e11);
  \draw [dashed, -latex, blue] (CNN1) to [bend right] (e12);
    \node[draw, single arrow,
              minimum height=3mm, minimum width=1mm,
              single arrow head extend=0.5mm,
              ] at (3.1,-1) {};
    \node[draw, single arrow,
              minimum height=3mm, minimum width=1mm,
              single arrow head extend=0.5mm,
              ] at (7.65,-1) {};
    \node[draw, single arrow,
              minimum height=3mm, minimum width=1mm,
              single arrow head extend=0.5mm,
              ] at (12.2,-1) {};
  \begin{pgfonlayer}{background}
    \filldraw [line width=4mm,join=round,black!10]
      (image1.north  -| a1.west)  rectangle (CNNt1.south  -| CNNt1.east)
      (e1.north  -| e1.west)  rectangle (CNNt2.south  -| CNNt2.east)
      (f1.north  -| f1.west)  rectangle (CNNt3.south  -| CNNt3.east)
      (g1.north  -| g1.west)  rectangle (CNNt3.south  -| CNNt4.east);
  \end{pgfonlayer}
\end{tikzpicture}
\caption{Overall diagram for our approach}
\label{fig:overall}
\end{figure*}




An overview of our approach is shown in Fig.\ref{fig:overall}. First, we build two graphs from the knowledge base and the CNN(Fig.\ref{fig:overall}(a-b)). We then apply entity alignment by aligning the labelled nodes across these two graphs (Fig.\ref{fig:overall}(c)) that have the same meaning. Finally, in the embedding space generated by the entity alignment stage, we explore the knowledge graph nodes which are the nearest neighbours of the CNN feature nodes with the goal of labelling the CNN feature nodes, based on equivalent nodes aligned from the knowledge graph. The following sections will introduce the details of each step.

\subsection{Building a graph from a knowledge base}
Knowledge bases are mainly stored as two kinds of triples: \textit{(head, relationship, tail)} and \textit{(entity, property, value)}. The first one represents the relationships between entities, such as \textit{(leopard, IsA, cat)} \cite{speer2017conceptnet}. An example of the second kind is \textit{(Texas, areaTotal, ``696241.0”)} \cite{sun2017cross}. 
For our purposes, we use the first kind of triple, which describes relationships explicitly. 
Triples can have associated weights which represents the \textit{strength} of the relationship between the head and the tail entities. 
To build a graph from the knowledge base, entities in the knowledge base become nodes in the graph.
Since the knowledge base is very large as we are using ConceptNet, we will use a subset. The starting points of building the knowledge graph are the entities from the knowledge base that have the same meaning as the CNN outputs (i.e. classes). Using a breadth-first search  strategy through the knowledge base, we then add in other entities that are directly connected to these entities(i.e. one-hop) that have a specific relationship and weight. These knowledge graph outputs nodes are connected to the nodes from the  knowledge base that had the largest triple relationship weights. 
There are many kinds of triple relationships (such as IsA, HasA). In an ideal knowledge base, we would aim to include only those triples that are relevant to visual features. As we do not have this kind of knowledge base, we use ConceptNet \cite{speer2017conceptnet}, which is an existing knowledge base and adjust it by selecting relationships that are more likely to map to visual features. For instances, triple \textit{(A cat, HasA, a nose)} is more related to visual information than triple \textit{(a cat, Desires, meow)}. We have explored the results using ``IsA'' and ``HasA', but more experiments using different combinations of relationships can be done in the future work. 

\subsection{Building a graph from the CNN}
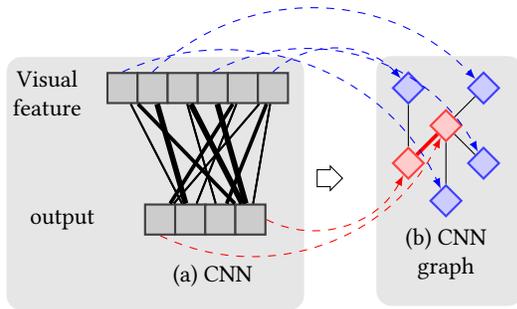
\begin{figure}[h]\centering
\begin{tikzpicture}[node distance=0.9cm,bend angle=45,auto,edge_style/.style={draw=black, ultra thick}]
  
  \tikzstyle{place}=[circle,thick,draw=blue!75,fill=blue!20,minimum size=3mm]
  \tikzstyle{red place}=[place,draw=red!75,fill=red!20]
  \tikzstyle{cnn}=[diamond,thick,draw=blue!75,fill=blue!20,minimum size=3mm]
  \tikzstyle{red cnn}=[cnn,draw=red!75,fill=red!20]
  \tikzstyle{highlight}=[circle,thick,draw=black,minimum size=8mm]
  \tikzstyle{transition}=[rectangle,thick,draw=black!75,
  			  fill=black!20,minimum size=4mm,node distance=0.4cm]

 \begin{scope}[xshift=1.5cm]
    \node [transition] at (0, 0) (a1) {};
    \node [transition] (a2) [right of=a1]                     {};
    \node [transition] (a3) [right of=a2]                     {};
    \node [transition] (a4) [right of=a3]                     {};
    \node [transition] (a5) [right of=a4]                     {};
    \node [transition] (a6) [right of=a5]                     {};
    
    \node [transition] at (0.5, -1.75) (b1) {};
    \node [transition] (b2) [right of=b1]                     {};
    \node [transition] (b3) [right of=b2]                     {};
    \node [transition] (b4) [right of=b3]                     {};

    \draw [line width=0.25mm](a1) -- (b2);
    \draw [line width=0.5mm](a1) -- (b4);
    \draw [line width=0.75mm](a2) -- (b2);
    \draw [line width=0.25mm](a5) -- (b2);
    \draw [line width=0.5mm](a5) -- (b1);
    \draw [line width=0.25mm](a3) -- (b3);
    \draw [line width=0.75mm](a3) -- (b4);
    \draw [line width=0.5mm](a6) -- (b3);
    \draw [line width=0.25mm](a6) -- (b1);
    \draw [line width=0.75mm](a4) -- (b4);
    \draw [line width=0.25mm](a6) -- (b4);
    \draw [line width=0.25mm](a1) -- (b2);
    \draw [line width=0.25mm](a1) -- (b4);
    \draw [line width=0.25mm](a2) -- (b2);
    \draw [line width=0.25mm](a5) -- (b2);
    \draw [line width=0.25mm](a5) -- (b1);
    \draw [line width=0.25mm](a3) -- (b3);
    \draw [line width=0.25mm](a3) -- (b4);
    \draw [line width=0.25mm](a6) -- (b3);
    \draw [line width=0.25mm](a6) -- (b1);
    \draw [line width=0.25mm](a4) -- (b4);
    \draw [line width=0.25mm](a6) -- (b4);
    
    \node (feature) at (-1, -0.1) {\begin{tabular}{cc}Visual \\feature\end{tabular}};
    \node (output) at (-0.8, -1.75) {output};
    \node (CNN) at (1.2, -2.5) {(a) CNN};
  \end{scope}

  \begin{scope}[xshift=3.5cm]
    
    \node [cnn] at (2.8, 0) (e7)                                {};
    \node [red cnn] at (2.3, -0.5)(e8) []                    {};
    \node [cnn] at (2.8, -1.0)(e9)                       {};
    \node [red cnn] at (1.8, -1.0)(e10)                      {};
    \node [cnn] at (2.3, -1.5)(e11)                      {};
    \node [cnn] at (1.8, 0)(e12) []                    {};
    \draw (e7) -- (e8);
    \draw (e8) -- (e9);
    \draw [red, line width=0.5mm] (e8) -- (e10);
    \draw (e8) -- (e11);
    \draw (e10) -- (e12);
    
    \node (CNNt2) at (2.3, -2.2) {\begin{tabular}{cc}(b) CNN \\graph\end{tabular}};
  \end{scope}
\draw [dashed, -latex, blue] (a6.north) to [bend left] (e12);
\draw [dashed, -latex, red] (b1.south) to [bend right] (e8);
\draw [dashed, -latex, blue] (a1.north) to [bend left] (e11);
\draw [dashed, -latex, red] (b4.east) to [bend right] (e10);
\draw [dashed, -latex, blue] (a2.north) to [bend left] (e7);
\draw [dashed, -latex, blue] (a4.north) to [bend left] (e9);

    \node[draw, single arrow,
              minimum height=3mm, minimum width=1mm,
              single arrow head extend=0.5mm,
              ] at (4.2,-1.2) {};
  \begin{pgfonlayer}{background}
    \filldraw [line width=4mm,join=round,black!10]
      (a1.north  -| output.west)  rectangle (CNN.south  -| a6.east)
      (e12.north  -| e12.west)  rectangle (CNNt2.south  -| CNNt2.east);
  \end{pgfonlayer}
\end{tikzpicture}
\caption{Diagram for building a graph from CNNs}
\label{fig:cnn_build}
\end{figure}

The visualization work of  \citet{zeiler2014visualizing} and other researchers have already shown that in a CNN, the features in the earlier layers in the network (close to the input) are detecting lower-level features. In a deep CNN such as VGG-16, the features in CNN 
outline the contour of some part of the input image \cite{zeiler2014visualizing}. The work of Grad-CAM\cite{selvaraju2017grad} reveals that the area representing the shape of prediction is the weighted combination of these CNN features. 
These works show that the prediction is made according to the features that are detected. The purpose of this step is to generate a graph from CNN that contains the outputs and the features that will bring most contribution to the outputs (akin to components of the output).The features in the higher network layers (close to output) are  combinations of lower level features with different weights. The visual feature for building our CNN graph is from the flatten layer. The flatten layer is the highest layer containing convolutional kernels, and has the highest level features. After the flatten layer, the fully connected layer combines the features in the flatten layer to detect the target classes. However, information is also lost during this combination. Therefore, we use the flatten layer features to build the CNN graph as it is a balance point of selecting features that are between too low and too high.  We let each feature and each output be a node in the graph (Fig.\ref{fig:cnn_build}). We want to connect those most relevant feature nodes with output nodes, to build a graph that the outputs and the important features are connected with each other. When an image is pushed into CNN, some features are activated. We want to choose the important features among these activated features.
There are several methods that we can choose from to rank the importance of a feature for an output including Pearson correlation \cite{benesty2009pearson}, DeConvolution, Guided Backprop \cite{springenberg2014striving}, LRP and DeepSHAP. We select Pearson correlation because it is both intuitive and widely used for ranking associations.  
 Other methods can be tested in the future.

\textbf{Step 1: Connect output nodes}
In the CNN side, the only thing we know are the labels of outputs. In the knowledge graph, we can locate these output nodes (a "dog" output in CNN will map to a "dog" node in the knowledge graph). Therefore, 
the CNN output nodes are the anchors to find meanings of visual features. Before attempting to connect any CNN features to  CNN output nodes, we connect those CNN outputs to each other, where those equivalent output nodes in the knowledge graph are directly connected to each other.  This  helps the entity alignment process to locate at least these output nodes, and thus enable the process to converge. 

\textbf{Step 2: Connect output nodes with visual features}
We then want to connect the most important CNN feature nodes correctly to the CNN output nodes, by identifying the subset of features that are most activated for each class.  We iterate through all the images in the dataset that trains the CNN(e.g. ImageNet), and derive an average of the importance (correlation) of each feature per output class. Now we know for each output node which features are most important. For example, for all images containing dog, features detecting tails and legs may have a higher activation. For each class, We average the correlation for each feature through the whole ImageNet dataset for that class. So, in the case of output node 'dog', visual feature nodes related to tails and legs likely be connected to that output node after averaging the feature correlations.  Fig.\ref{fig:cnn_build} shows a diagram of this process, simplified to show a subset of feature and output nodes. In Fig.\ref{fig:cnn_build}(a), the width of each line represents the strength of correlation of a feature to an output node. On the right side, the red nodes are the output nodes, and the blue ones are the CNN features.

\subsection{Entity Alignment: aligning the CNN and knowledge graphs}
Now we have our CNN graph and the knowledge graph. The process in the first two steps of Fig.\ref{fig:overall} is finished. The CNN graph contains the output nodes (one per label) and the most important features nodes associated with each output node. The knowledge graph contains the equivalent output nodes as the CNN, having been specifically selected from the knowledge base. 
The entity alignment algorithm will take the two graphs as input and generate embeddings for each nodes, so that all the nodes in the two graphs will be represented in the same embedding space. Similar to word embeddings \cite{le2014distributed}, the distance in the embedding space between each pair of nodes reflects the semantic distance between them. 
For a node in a graph, and that node is represented in the embedding space, the process of finding the corresponding node from the second graph with similar meaning is to look for the closest neighbour in the embedding space. 

To align two graphs successfully,  their \textit{structure} must be similar. However, the number of edges in the knowledge graph is significantly smaller than those in the CNN graph. Therefore, we have to select a subset of the edges in the CNN graph to match the number in the knowledge graph. There are two ways we can approach this: (1) For a knowledge graph that has $k$ edges, we sort all the correlations between visual features and output nodes in CNN, and take the top $k$ correlations with the feature nodes into account. (2) For a CNN output node $node_i$, there are $k_i$ visual features. We select the same number of $k_i$ edges of $node_i$ in the knowledge graph in order to build two graphs with similar structure. 

Correlations may be positive or negative. Negative gradients \cite{selvaraju2017grad} and correlations indicate that a particular feature is not predictive of a current output node. Thus, we only consider positive correlations, by matching the edge number($k_i$) of knowledge graph.
 
In order to create the embedding space during the alignment process, we need to train GCN-Align with matching pairs of CNN/ knowledge graph output nodes. The outputs classes (from the ImageNet dataset) in CNN form our CNN nodes, and the equivalent concepts in Concept net form our knowledge graph nodes, to provide us with 1000 node pairs. We select 90\% of the pairs randomly as the training set. The remaining 10\% of the pairs will be the test set. The training process is to minimize the distance between these 90\% pairs of nodes. The testing process is to check among those 10\% pairs of nodes to test whether in the generated embedding space, the output node is the closest to the equivalent node in the knowledge base.  


\subsection{Evaluation approach} 

To explore our hypotheses in the introduction session, we examine the embedding space and evaluate our feature labelling system in two ways.

\textbf{Alignment performance}
Firstly,  we evaluate the entity alignment process by measuring whether we can align the equivalent output nodes across the knowledge graph and the CNN graph. 
We use \textbf{Hit@1} score to show the performance of entity alignment, as used in other papers \cite{wang2018cross,xu2019cross}. In the embedding space, we select the nearest knowledge graph node to each CNN output class node in the test set, based on Euclidean distance. The Hit@1 measures the proportion of the CNN output nodes in the test set whose closest neighbour is the corresponding knowledge graph node with the same meaning. 

\textbf{Hit@k-nns chart}
Secondly, once we have established via the Hit@1 score that equivalent output nodes (from the CNN and knowledge graphs) are located together, the \textit{neighbourhood} of a Knowledge Graph node should be fully explored. We will use the closest knowledge base node to a CNN feature node to label the CNN feature node. 
The feature labelling system is the embedding space generated by the previous aligning step. 
A smaller distance between them means the visual feature is more likely to detect that meaning. We introduce the Hit@k-nns Chart. Here an independent image test dataset is used, which is independent of the training set used to train the CNN. The target classes i.e. output nodes are different from the dataset we used for training in Entity Alignment. For each test image, Hit@k-nns Chart will explore $k$ nearest visual feature neighbour nodes of the label of that image. The level of activated visual nodes in these neighbours will reflect how well the embedding space has captured the semantic meaning of visual features.
Here is the steps of plotting Hit@k-nns Chart:
\begin{itemize}
    \item A test image is passed through the pre-trained CNN.
    \item Feature nodes that are activated for that image will be identified.
    \item The set of k feature nodes that are the nearest neighbours of the target output knowledge graph node for the test image will be identified.
    \item The percentage of nearest neighbour feature nodes that are activated will be measured. (i.e. Hit@k-NN score)
    \item Both Hit@k-NN for a single image as k increases, or aggregated over a set of test images, is plotted.
\end{itemize}

The Hit@k-NN score of a test image will be plotted in a graph where the x-axis is the value $k$, which is the number of explored CNN feature nodes that are nearest neighbours. The y-axis is the percentage of activated nodes. 
In a perfect situation (as shown in Fig.\ref{fig:hitk}), all these nearest CNN feature nodes should be activated. The shape of the curve should be high for low values of k and dropping down as k increases. The perfect shape is drawn for the situation that assuming that all the closest CNN feature nodes are activated. All other not activated is not near the target output knowledge graph node.

\textbf{Embedding space visualisation}
We also visualise the generated embedding space to clearly show the placement of the nodes and the distances between CNN feature nodes and knowledge graph nodes by mapping it to a two-dimensional space using t-SNE \cite{maaten2008visualizing}.  We integrated Deconvolution \cite{simonyan2013deep} into the visualisation so that we display the contour that the visual feature node is detecting. 
An example of this can be seen in Figure~\ref{fig:isa_feature_visual_tv}. In the middle it is the t-SNE mapped embedding space. For each visual feature, there is the visualisation in the grey box to the right of the test images. 


\subsection{Experiment details
}

\begin{table}[]
\begin{tabular}{|l|l|}
\hline
CNN backbone            & VGG-16(ImageNet)   \\ \hline
Importance measurement   & Pearson Correlation \\ \hline
Hit@k chart test set    & COCO dataset       \\ \hline
Knowledge Base          & ConceptNet 5.5     \\ \hline
Entity Alignment method & GCN-Align          \\ \hline
\end{tabular}
\caption{Experiment details}
\label{tab:experimentdetail}
\end{table}

This section outlines the implementation details of our experimental evaluation. 

We use the VGG-16 \cite{simonyan2014very} model as our CNN. The version of VGG-16 that we used is trained on the ImageNet dataset.  In VGG-16 the flatten layer is a 25088 dimensional vector and output is a 1000 dimensional vector, as the VGG-16 used here is trained to distinguish between 1000 classes. 
We use the training set of ImageNet \cite{deng2009imagenet} to calculate the Pearson Correlation between image features and output nodes to select important features from the flatten layer for our CNN graph. 
In the entity alignment process, typically a large number of nodes - up to 100,00 - are used for training the alignment algorithm. 
\cite{sun2017cross,wang2018cross,zhu2020collective}.
CNNs, however, do not typically detect enough categories to support this scale of entity alignment training. In our case, the entity alignment process is driven by the number of CNN outputs nodes. The 1000 categories in the ImageNet datasets form the combined testing/ training set in GCN-Align training process. To further \textit{test} the embedding space generated by GCN-Align, we use a separate image dataset, the COCO dataset \cite{lin2014microsoft}.
We identified test images from COCO whose labels do not occur in the training images in ImageNet and these images were used for testing. 

The knowledge base we used for building the knowledge graph is ConceptNet 5.5 \cite{speer2017conceptnet}.  We identified entities in ConceptNet that were exactly the same as the 1000 categories of the ImageNet dataset. These output nodes became the nodes of the knowledge graph built from ConceptNet.  We also identified an additional 62 categories from COCO which were the different from the ImageNet categories and these entities were included as nodes in the knowledge graph to allow us to use them at a later stage for testing.  
We selected the  \textit{``IsA"} and \textit{``HasA"} relationships, as we expect that these relationships are more likely to be between visual features. 
We suggest that the \textit{``IsA"} relationship has the potential of an inference relationship between nodes and may be useful for zero-shot Learning. The ``HasA" relationship could help us find the components of an object which  may be useful in explanation.
We built two knowledge graphs for our evaluation, the first from the \textit{``IsA"} relationships with the output nodes, the second from the  \textit{``HasA"} relationships.

For entity alignment, we used GCN-Align \cite{wang2018cross}. This method is purely based on the graph structure, which is suitable for our purposes. The input to GCN-align does not need any word embeddings or attribute embeddings. Table.\ref{tab:experimentdetail} shows details of experiments.


\section{Results and discussion}

\subsection{Alignment performance}

\begin{figure}[h]
\centering
    \begin{subfigure}[b]{.23\textwidth}
    \includegraphics[width=\textwidth]{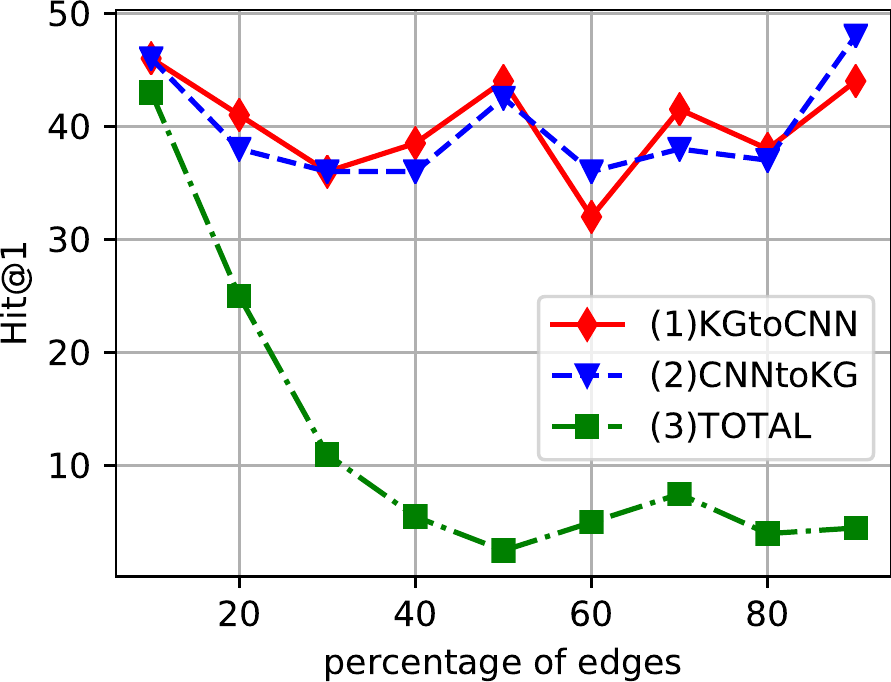}  
    \caption{Hit@1 score of using the \textit{``IsA"} relationship. }
    \label{fig:isa-hit1}
    \end{subfigure}
    \begin{subfigure}[b]{.23\textwidth}
    \includegraphics[width=\textwidth]{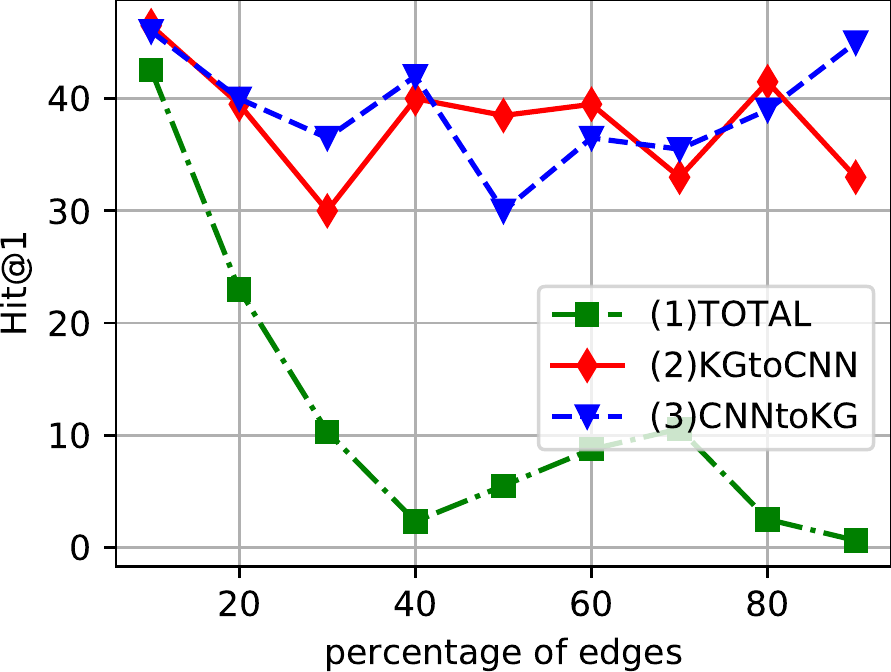}  
    \caption{Hit@1 score of using the \textit{``HasA"} relationship}
    \label{fig:hasa-hit1}
    \end{subfigure}
    \vspace{-2mm}
    \caption{Hit@1 score comparison for three different graph building approaches. (1) KG-to-CNN: Letting the knowledge graph match the edge number of each node in the CNN graph. (2) CNN-to-KG: Letting the CNN graph match the edge number of each node in the knowledge graph. (3) TOTAL: Do not match the edge number of each node, and just match the total number of edges in two graphs}
    \label{fig:hit1}
\end{figure}

Hit@1 score measures whether the closest neighbour of a CNN output node is the corresponding knowledge graph node with the same meaning. Fig.\ref{fig:hit1} shows the align scores (Hit@1) based on three different approaches to building our two graphs, depending on how the edges are added between nodes. The x-axis is the percentage of total edges selected from the knowledge base and the y-axis is the value of Hit@1 score. Edges are added 10\% at a time in all cases. This score measures the proportion  of test set nodes from one graph that are the closest node to the matching node from the other graph in the embedding space. 

Here are the three ways of adding edges:
\begin{itemize}
    \item \textbf{KGtoCNN} starts with the nodes in the knowledge graph and at each iteration adds 10\% of the total number of KG edges to each CNN node. At the same time, two corresponding nodes with same meaning in two graphs have the same edge number.
    \item \textbf{CNNtoKG} does the opposite starting with the CNN and at each iteration adding 10\% of the total number of CNN edges to each KG node. Also, two corresponding nodes with same meaning in two graphs have the same edge number.
    \item \textbf{TOTAL} starts with the nodes in the knowledge graph and at each iteration adds 10\% of the total number of KG edges to each CNN node, too. But TOTAL does not keep the edge number of each pair of nodes same.
\end{itemize}

In Fig.\ref{fig:isa-hit1} (3) TOTAL shows how the Hit@1 score changes when we add the same number of edges over the 1000 categories each time when building our CNN graph and knowledge graphs (and where the knowledge graph is derived from extracting the is-a relationships from ConceptNet). The Hit@1 score for adding 10\% of total number of edges is 43\% which is the average performance achieved by GCN-Align generally\cite{wang2018cross}. We observe that the performance is dropping as we add more edges. Hit@1 score becomes lower than 10\% after we add 40\% of the edges. (1)KG-to-CNN has a different strategy of adding edges.We still add 10\% of the total edges each time. However, across all those 1000 output nodes, we let the number of edges from the knowledge graph node match the number of edges in the corresponding node in CNN. For example, at the step of 10\%, on the CNN side, we select the top 28994 edges among all the 289944 edges. For example, the output node ``bench" may have $k$ edges in this 28994 edges. On the knowledge graph side, we sort all the edges of ``bench" in descending order and select the top $k$ edges. If in the knowledge graph ``bench" does not have enough edges, we will select all of them. This process ensures that every output node will have almost the same number of edges. The Hit@1 score for KGtoCNN is more stable compared to (3)TOTAL. Unlike (1)KG-to-CNN, which is making knowledge graph match CNN graph, (2)CNN-to-KG is letting the CNN match the knowledge graph. The alignment score performance is similar. 

Fig.\ref{fig:hasa-hit1} shows the Hit@1 score of aligning the CNN graph with the knowledge graph, but here the knowledge graph is created from the ``HasA" relationships in ConceptNet. Compared to the graph generated from ``IsA" relationships, the graph created from ``HasA" relationships is more sparse. In total there are 13774 edges and 1263 nodes.
 The figure shows that entity alignment performance as the number of edges increase is very similar to that of the ``IsA" relationships across the three different approaches to building the graphs -  although  the performance using a ``HasA" graph is slightly lower than using an ``IsA" graph.  This suggests that the CNN graph can be aligned  with knowledge base graphs that are generated from different kinds of relationships.

\subsection{Hit@k-nns chart}

\begin{figure}[h]
\begin{subfigure}{.23\textwidth}
  \includegraphics[width=\linewidth]{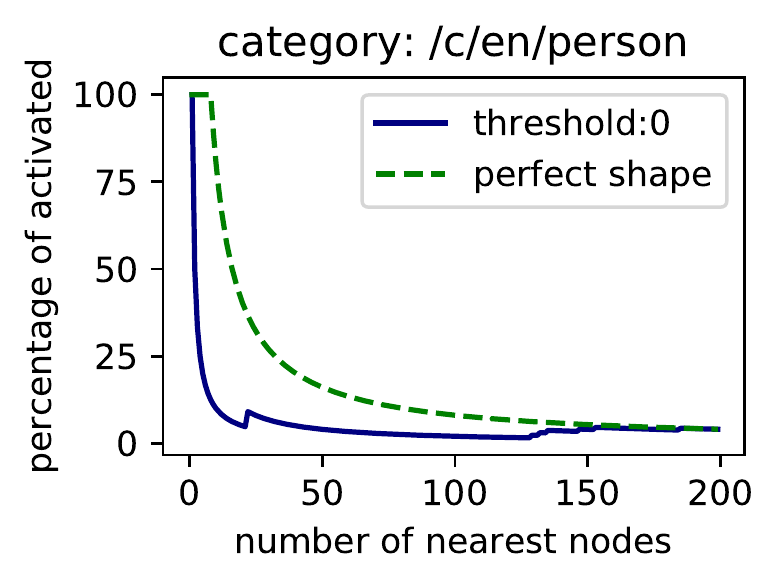}  
  \caption{Person(\textit{``IsA"})}
  \label{fig:hitk-isa-person-sub}
\end{subfigure}
\begin{subfigure}{.23\textwidth}

  \includegraphics[width=\linewidth]{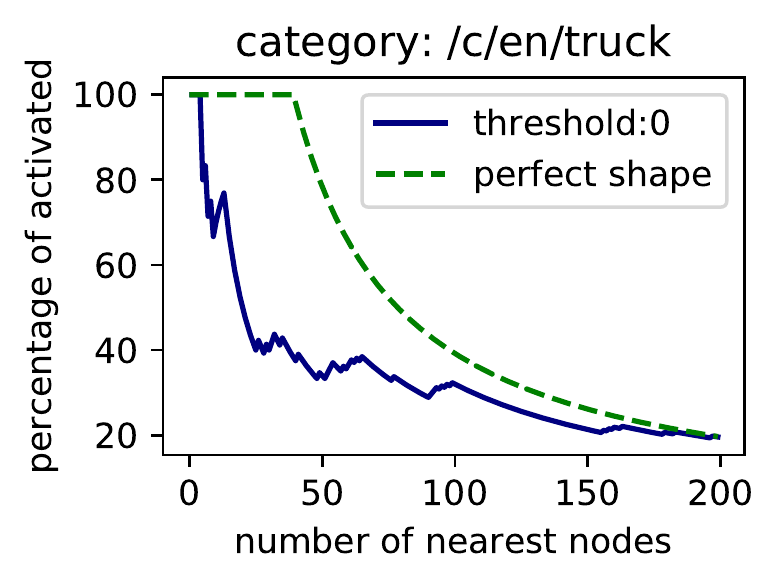}  
  \caption{Truck(\textit{``IsA"})}
  \label{fig:hitk-isa-truck-sub}
\end{subfigure}
\begin{subfigure}{.23\textwidth}
  \includegraphics[width=\linewidth]{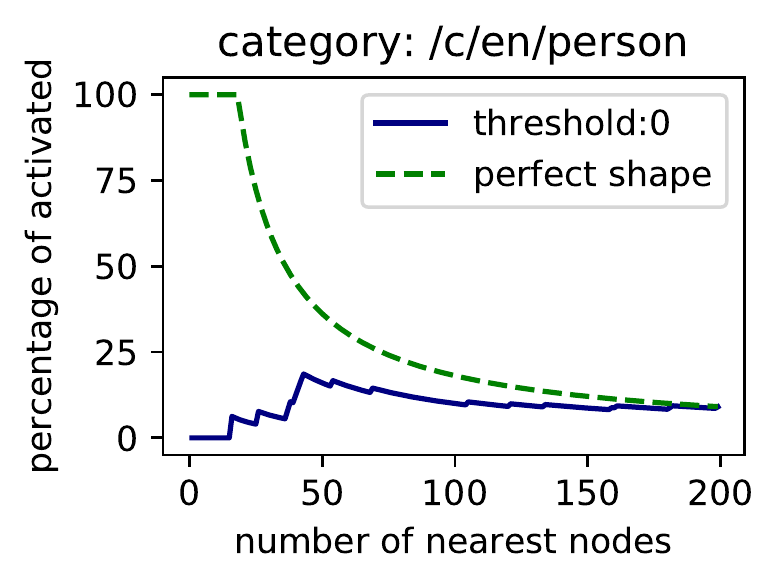}  
  \caption{Person(\textit{``HasA"})}
  \label{fig:hitk-hasa-person-sub}
\end{subfigure}
\begin{subfigure}{.23\textwidth}
  \includegraphics[width=\linewidth]{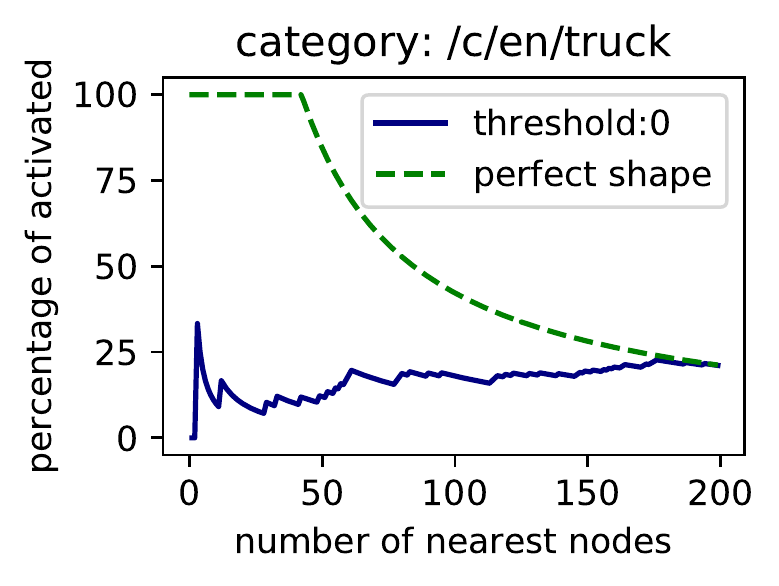}  
  \caption{Truck(\textit{``HasA"})}
  \label{fig:hitk-hasa-truck-sub}
\end{subfigure}
\centering
\begin{subfigure}{.18\textwidth}
  \includegraphics[width=\linewidth]{example_image.jpg}  
  \caption{The example image}
  \label{fig:hitk-image}
\end{subfigure}
\vspace{-4mm}
\caption{Hit@k-nns charts of the example image. It has the label of ``person" and ``truck".}
\label{fig:hitk}
\end{figure}

As we have a reasonable alignment performance for entity alignment based on the Hit@1 score, we can confirm that the graph alignment process places output nodes in our two graphs with the same semantic meaning very close to each other in an embedding space. Now we explore this embedding space to see whether it places CNN feature nodes close to the associated knowledge base nodes. To assess this, we used the embedding spaces generated by (1)KGtoCNN in Fig.\ref{fig:hit1} with 50\% of edges. 

The COCO dataset is used to evaluate the embedding space generated from the graph alignment process. The VGG-16 CNN has never been exposed to the COCO dataset so it provides us with an independent set of test images.

Fig.\ref{fig:hitk} is an example of Hit@k-nns chart for an image from COCO validation set. Fig.\ref{fig:hitk-image} shows the image in question. It has the labels of ``truck" and ``person", both of which can be seen in the image.  Neither of these two labels are in the ImageNet dataset. 
We identified the 200 nearest CNN feature nodes around the ``truck" and ``person" node from the knowledge graph. The graph represents the percentage of  CNN feature nodes within that 200 nodes with an activation larger than a threshold of 0 to exclude negative correlations. 


In Fig.\ref{fig:hitk} the green dot line is the Hit@k-NN chart is the  theoretically perfect situation that all the activated CNN feature nodes are the nearest neighbours of the label node, using a threshold of 0. 
Fig.\ref{fig:hitk-isa-person-sub} shows the embedding space generated only using ``IsA" relationship, it shows that a high proportion of the nearest neighbours of the node ``person" are activated CNN feature nodes.

Fig.\ref{fig:hitk-isa-truck-sub} shows that at $k=1$ all CNN features that are closest to the truck node are all activated. Although the proportion of activated nodes drops as $k$ increases, this does not necessarily suggest poor performance as all the visual feature nodes will not necessarily need  to be activated to detect the content in an image. 

For the other embedding space generated by ``HasA", the result is not as good. Fig.\ref{fig:hitk-hasa-person-sub} and Fig.\ref{fig:hitk-hasa-truck-sub} show that the Hit@k-nns line is far from the perfect shape. This shows when using the ``HasA" relationship to build the knowledge graph that the nearest CNN feature nodes surrounding ``person" and ``truck" do not directly relate to these visual concepts.


\subsection{Embedding space visualization}
    

\begin{figure*}[h]
    \centering
    \includegraphics[width=.8\linewidth]{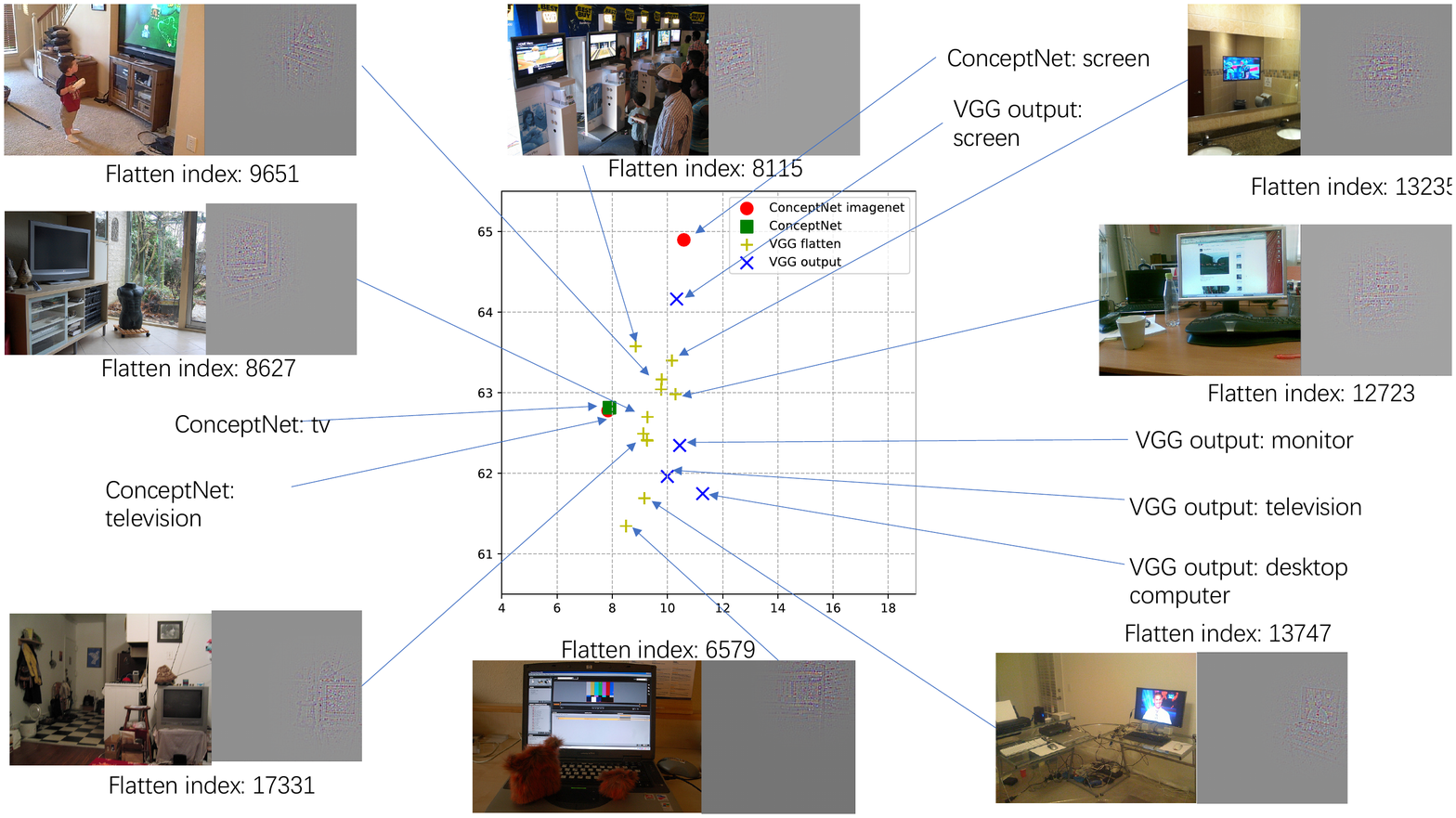}  
    \vspace{-14mm}
    \caption{An example of t-SNE mapped embedding space generated only using \textit{``IsA"}}
    \label{fig:isa_feature_visual_tv}

\end{figure*}

\begin{figure*}[h]
    \centering
    \includegraphics[width=.8\linewidth]{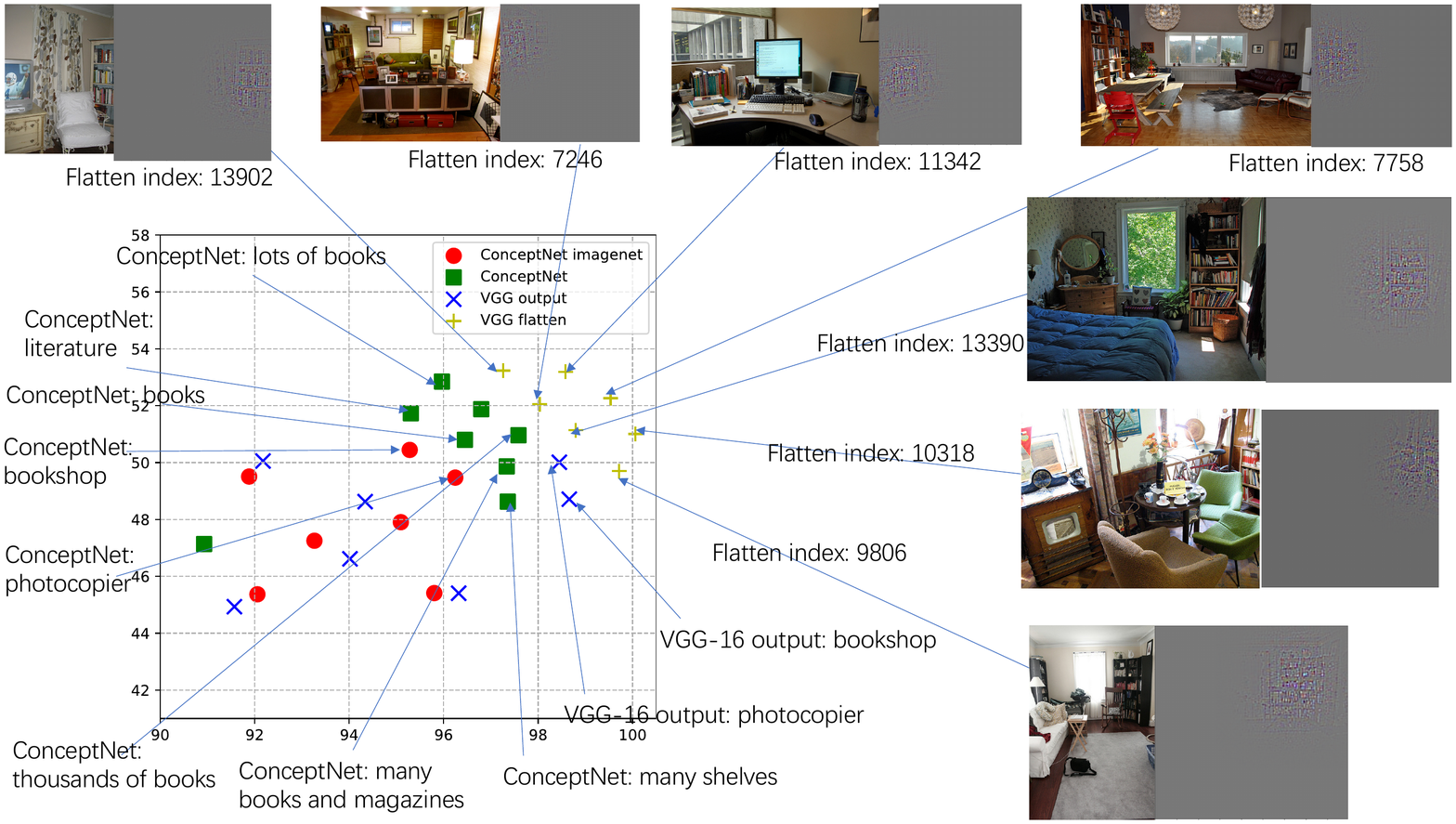}  
    \vspace{-12mm}
    \caption{An example of t-SNE mapped embedding space generated only using \textit{``HasA"}}
    \label{fig:Hasa_feature_visual}

\end{figure*}

Since the feature labelling system we proposed uses GCN-Align to generate the embedding space, it produces a 200-dimensional vector for describing the location of each node i.e. the node embedding. To visualize the embedding space, we applied t-SNE\cite{maaten2008visualizing}  it and mapped the node embedding to a 2-dimensional space. Fig.\ref{fig:isa_feature_visual_tv} is a section of the visualisation of the nearest nodes in the embedding space generated from the ``IsA" for the test image shown in Fig.\ref{fig:hitk}.  The nodes shown in this visualisation are all those that are found in the section of the embedding space represented in the visualisation. No nodes have been removed. 
In this figure the red circles are the ConceptNet output nodes. The yellow ``+" are the VGG-16 network output nodes. 
The blue ``$\times$" symbols represent the visual CNN feature nodes in the flatten layer. The red square symbols are the ConceptNet nodes that are connected to the output nodes. For each CNN feature node in flatten layer, we select the test image from the COCO dataset that has the highest activation of that feature. 
In addition to visualising the embedding space we use Deconvolution to visualise the CNN feature which is displayed beside the test image (the grey box) and shows the focus of the CNN feature for that image. 

In Fig.\ref{fig:isa_feature_visual_tv} we can see that nodes with similar semantic meanings are located close to each other, for example, the ``screen" node network output node from VGG-16 and the ``screen" node from ConceptNet.  Nodes that have similar meanings, such as ``tv" and television from ConceptNet, are positioned almost at the same point. The closest network output features all relate to objects that are similar semantically, ``monitor", ``television", ``desktop computer", ``screen".   
For each of these visual features, the Deconvolution visualization shows that the content they are detecting matches the meanings of these nodes. 
This visualisation shows that for  the embedding space generated for the graph alignment of ConceptNet using the ``IsA" relationship captures this ``IsA" relationship with all nodes in this version of the embedding space related to similar objects.

Fig.\ref{fig:Hasa_feature_visual} is an example of the embedding space generated from a ConceptNet graph using the ``HasA" relationship. 
In this figure, features detecting books are also near to the nodes from both VGG-16 and ConceptNet. Moreover, in comparison to Fig.\ref{fig:isa_feature_visual_tv},  there are significantly more ConceptNet nodes in this neighbourhood, the green squares which are not the output nodes that the graphs were aligned on.  
These nodes are potentially new categories and the embedding space has located them close to relevant visual features. 
The visualization shows that our approach can place visual features in a semantic embedding space and the meanings of nearest neighbour nodes could be a reference for naming the visual feature.


\section{Conclusion}

Current research works can explain visual features with several external meanings, but the feature labelling system is limited because the concepts for explaining visual features are selected manually. In this paper, we have presented an approach, based on entity alignment,  to  finding the meanings of visual features in CNNs using all the related concepts selected automatically from a large knowledge base, ConceptNet. We built a graph from VGG-16 by selecting the important features, and applied entity alignment between the sub-graph of ConceptNet and the VGG-16 graph to match the visual features with semantic meanings. The entity alignment reached an average level Hit@1 score of GCN-Align. The Hit@k-nns charts shows that the true label of an image is surrounded by activated visual features. The visualization of visual features with t-SNE mapping of embedding space shows that the visual features are close to the concepts that they are detecting, and those concepts will explain the meaning of that visual feature.  

The main limitation of our work at this point is the knowledge base is too noisy. It is full of non-visual concepts, such as (a cat, Desires, meow). We know it may be irrelevant because it is describing a sound. In other cases, however, it is hard to tell whether the triple is visual or not. These non-visual triples will interfere with the alignment process. Another potential limitation is that the visual feature may not have the same meaning for all the time. A same visual feature may be detecting the screen of a monitor in one image but it may be still detecting the frame of a window in another image. 

Having shown the promise of enriching CNNs with external knowledge bases via entity alignment, the next phase of our  work will involve the following (1) Acquiring a more suitable knowledge source. The current external knowledge source, ConceptNet, is built from a large language corpus. It was not created for the purpose of labelling or analysing visual features, so the noise in the knowledge base will perturb the feature labelling system. (2) Evaluating additional methods for finding the most important features/nodes. (3) Using the embedding space to reason for unseen classes for Zero-shot learning based on feature matching. (4) Combining with object detection: After doing reasoning using the embedding space, we could follow the existing methods \cite{selvaraju2017grad} to locate the area of an object. 
    

\begin{acks}
This work was funded by Science Foundation Ireland through the SFI Centre for Research Training in Machine Learning (18/CRT/6183)
\end{acks}

\bibliographystyle{ACM-Reference-Format}

\begin{thebibliography}{44}

\ifx \showCODEN    \undefined \def \showCODEN     #1{\unskip}     \fi
\ifx \showDOI      \undefined \def \showDOI       #1{#1}\fi
\ifx \showISBNx    \undefined \def \showISBNx     #1{\unskip}     \fi
\ifx \showISBNxiii \undefined \def \showISBNxiii  #1{\unskip}     \fi
\ifx \showISSN     \undefined \def \showISSN      #1{\unskip}     \fi
\ifx \showLCCN     \undefined \def \showLCCN      #1{\unskip}     \fi
\ifx \shownote     \undefined \def \shownote      #1{#1}          \fi
\ifx \showarticletitle \undefined \def \showarticletitle #1{#1}   \fi
\ifx \showURL      \undefined \def \showURL       {\relax}        \fi
\providecommand\bibfield[2]{#2}
\providecommand\bibinfo[2]{#2}
\providecommand\natexlab[1]{#1}
\providecommand\showeprint[2][]{arXiv:#2}

\bibitem[\protect\citeauthoryear{Akbari, Fathian, and Badie}{Akbari
  et~al\mbox{.}}{2009}]%
        {akbari2009improved}
\bibfield{author}{\bibinfo{person}{Ismail Akbari}, \bibinfo{person}{Mohammad
  Fathian}, {and} \bibinfo{person}{Kambiz Badie}.}
  \bibinfo{year}{2009}\natexlab{}.
\newblock \showarticletitle{An improved mlma+ and its application in ontology
  matching}. In \bibinfo{booktitle}{\emph{2009 Innovative Technologies in
  Intelligent Systems and Industrial Applications}}. IEEE,
  \bibinfo{pages}{56--60}.
\newblock


\bibitem[\protect\citeauthoryear{Arrieta, D{\'\i}az-Rodr{\'\i}guez, Del~Ser,
  Bennetot, Tabik, Barbado, Garc{\'\i}a, Gil-L{\'o}pez, Molina, Benjamins,
  et~al\mbox{.}}{Arrieta et~al\mbox{.}}{2020}]%
        {arrieta2020explainable}
\bibfield{author}{\bibinfo{person}{Alejandro~Barredo Arrieta},
  \bibinfo{person}{Natalia D{\'\i}az-Rodr{\'\i}guez}, \bibinfo{person}{Javier
  Del~Ser}, \bibinfo{person}{Adrien Bennetot}, \bibinfo{person}{Siham Tabik},
  \bibinfo{person}{Alberto Barbado}, \bibinfo{person}{Salvador Garc{\'\i}a},
  \bibinfo{person}{Sergio Gil-L{\'o}pez}, \bibinfo{person}{Daniel Molina},
  \bibinfo{person}{Richard Benjamins}, {et~al\mbox{.}}}
  \bibinfo{year}{2020}\natexlab{}.
\newblock \showarticletitle{Explainable Artificial Intelligence (XAI):
  Concepts, taxonomies, opportunities and challenges toward responsible AI}.
\newblock \bibinfo{journal}{\emph{Information Fusion}}  \bibinfo{volume}{58}
  (\bibinfo{year}{2020}), \bibinfo{pages}{82--115}.
\newblock


\bibitem[\protect\citeauthoryear{Bach, Binder, Montavon, Klauschen, M{\"u}ller,
  and Samek}{Bach et~al\mbox{.}}{2015}]%
        {bach2015pixel}
\bibfield{author}{\bibinfo{person}{Sebastian Bach}, \bibinfo{person}{Alexander
  Binder}, \bibinfo{person}{Gr{\'e}goire Montavon}, \bibinfo{person}{Frederick
  Klauschen}, \bibinfo{person}{Klaus-Robert M{\"u}ller}, {and}
  \bibinfo{person}{Wojciech Samek}.} \bibinfo{year}{2015}\natexlab{}.
\newblock \showarticletitle{On pixel-wise explanations for non-linear
  classifier decisions by layer-wise relevance propagation}.
\newblock \bibinfo{journal}{\emph{PloS one}} \bibinfo{volume}{10},
  \bibinfo{number}{7} (\bibinfo{year}{2015}), \bibinfo{pages}{e0130140}.
\newblock


\bibitem[\protect\citeauthoryear{Bau, Zhou, Khosla, Oliva, and Torralba}{Bau
  et~al\mbox{.}}{2017}]%
        {bau2017network}
\bibfield{author}{\bibinfo{person}{David Bau}, \bibinfo{person}{Bolei Zhou},
  \bibinfo{person}{Aditya Khosla}, \bibinfo{person}{Aude Oliva}, {and}
  \bibinfo{person}{Antonio Torralba}.} \bibinfo{year}{2017}\natexlab{}.
\newblock \showarticletitle{Network dissection: Quantifying interpretability of
  deep visual representations}. In \bibinfo{booktitle}{\emph{Proceedings of the
  IEEE conference on computer vision and pattern recognition}}.
  \bibinfo{pages}{6541--6549}.
\newblock


\bibitem[\protect\citeauthoryear{Benesty, Chen, Huang, and Cohen}{Benesty
  et~al\mbox{.}}{2009}]%
        {benesty2009pearson}
\bibfield{author}{\bibinfo{person}{Jacob Benesty}, \bibinfo{person}{Jingdong
  Chen}, \bibinfo{person}{Yiteng Huang}, {and} \bibinfo{person}{Israel Cohen}.}
  \bibinfo{year}{2009}\natexlab{}.
\newblock \showarticletitle{Pearson correlation coefficient}.
\newblock In \bibinfo{booktitle}{\emph{Noise reduction in speech processing}}.
  \bibinfo{publisher}{Springer}, \bibinfo{pages}{1--4}.
\newblock


\bibitem[\protect\citeauthoryear{Bordes, Usunier, Garcia-Duran, Weston, and
  Yakhnenko}{Bordes et~al\mbox{.}}{2013}]%
        {bordes2013translating}
\bibfield{author}{\bibinfo{person}{Antoine Bordes}, \bibinfo{person}{Nicolas
  Usunier}, \bibinfo{person}{Alberto Garcia-Duran}, \bibinfo{person}{Jason
  Weston}, {and} \bibinfo{person}{Oksana Yakhnenko}.}
  \bibinfo{year}{2013}\natexlab{}.
\newblock \showarticletitle{Translating embeddings for modeling
  multi-relational data}. In \bibinfo{booktitle}{\emph{Advances in neural
  information processing systems}}. \bibinfo{pages}{2787--2795}.
\newblock


\bibitem[\protect\citeauthoryear{Carlson, Betteridge, Kisiel, Settles,
  Hruschka, and Mitchell}{Carlson et~al\mbox{.}}{2010}]%
        {carlson2010toward}
\bibfield{author}{\bibinfo{person}{Andrew Carlson}, \bibinfo{person}{Justin
  Betteridge}, \bibinfo{person}{Bryan Kisiel}, \bibinfo{person}{Burr Settles},
  \bibinfo{person}{Estevam~R Hruschka}, {and} \bibinfo{person}{Tom~M
  Mitchell}.} \bibinfo{year}{2010}\natexlab{}.
\newblock \showarticletitle{Toward an architecture for never-ending language
  learning}. In \bibinfo{booktitle}{\emph{Twenty-Fourth AAAI conference on
  artificial intelligence}}.
\newblock


\bibitem[\protect\citeauthoryear{Castelvecchi}{Castelvecchi}{2016}]%
        {castelvecchi2016can}
\bibfield{author}{\bibinfo{person}{Davide Castelvecchi}.}
  \bibinfo{year}{2016}\natexlab{}.
\newblock \showarticletitle{Can we open the black box of AI?}
\newblock \bibinfo{journal}{\emph{Nature News}} \bibinfo{volume}{538},
  \bibinfo{number}{7623} (\bibinfo{year}{2016}), \bibinfo{pages}{20}.
\newblock


\bibitem[\protect\citeauthoryear{Chen, Shrivastava, and Gupta}{Chen
  et~al\mbox{.}}{2013}]%
        {chen2013neil}
\bibfield{author}{\bibinfo{person}{Xinlei Chen}, \bibinfo{person}{Abhinav
  Shrivastava}, {and} \bibinfo{person}{Abhinav Gupta}.}
  \bibinfo{year}{2013}\natexlab{}.
\newblock \showarticletitle{Neil: Extracting visual knowledge from web data}.
  In \bibinfo{booktitle}{\emph{Proceedings of the IEEE international conference
  on computer vision}}. \bibinfo{pages}{1409--1416}.
\newblock


\bibitem[\protect\citeauthoryear{Deng, Dong, Socher, Li, Li, and Fei-Fei}{Deng
  et~al\mbox{.}}{2009}]%
        {deng2009imagenet}
\bibfield{author}{\bibinfo{person}{Jia Deng}, \bibinfo{person}{Wei Dong},
  \bibinfo{person}{Richard Socher}, \bibinfo{person}{Li-Jia Li},
  \bibinfo{person}{Kai Li}, {and} \bibinfo{person}{Li Fei-Fei}.}
  \bibinfo{year}{2009}\natexlab{}.
\newblock \showarticletitle{Imagenet: A large-scale hierarchical image
  database}. In \bibinfo{booktitle}{\emph{2009 IEEE conference on computer
  vision and pattern recognition}}. Ieee, \bibinfo{pages}{248--255}.
\newblock


\bibitem[\protect\citeauthoryear{Gao, Zhang, and Xu}{Gao et~al\mbox{.}}{2019}]%
        {gao2019know}
\bibfield{author}{\bibinfo{person}{Junyu Gao}, \bibinfo{person}{Tianzhu Zhang},
  {and} \bibinfo{person}{Changsheng Xu}.} \bibinfo{year}{2019}\natexlab{}.
\newblock \showarticletitle{I Know the Relationships: Zero-Shot Action
  Recognition via Two-Stream Graph Convolutional Networks and Knowledge
  Graphs}.
\newblock  (\bibinfo{year}{2019}).
\newblock


\bibitem[\protect\citeauthoryear{Gatys, Ecker, and Bethge}{Gatys
  et~al\mbox{.}}{2016}]%
        {gatys2016image}
\bibfield{author}{\bibinfo{person}{Leon~A Gatys}, \bibinfo{person}{Alexander~S
  Ecker}, {and} \bibinfo{person}{Matthias Bethge}.}
  \bibinfo{year}{2016}\natexlab{}.
\newblock \showarticletitle{Image style transfer using convolutional neural
  networks}. In \bibinfo{booktitle}{\emph{Proceedings of the IEEE conference on
  computer vision and pattern recognition}}. \bibinfo{pages}{2414--2423}.
\newblock


\bibitem[\protect\citeauthoryear{Gilpin, Bau, Yuan, Bajwa, Specter, and
  Kagal}{Gilpin et~al\mbox{.}}{2018}]%
        {gilpin2018explaining}
\bibfield{author}{\bibinfo{person}{Leilani~H Gilpin}, \bibinfo{person}{David
  Bau}, \bibinfo{person}{Ben~Z Yuan}, \bibinfo{person}{Ayesha Bajwa},
  \bibinfo{person}{Michael Specter}, {and} \bibinfo{person}{Lalana Kagal}.}
  \bibinfo{year}{2018}\natexlab{}.
\newblock \showarticletitle{Explaining explanations: An approach to evaluating
  interpretability of machine learning}.
\newblock \bibinfo{journal}{\emph{arXiv preprint arXiv:1806.00069}}
  (\bibinfo{year}{2018}).
\newblock


\bibitem[\protect\citeauthoryear{Hao, Zhang, He, Liu, and Zhao}{Hao
  et~al\mbox{.}}{2016}]%
        {hao2016joint}
\bibfield{author}{\bibinfo{person}{Yanchao Hao}, \bibinfo{person}{Yuanzhe
  Zhang}, \bibinfo{person}{Shizhu He}, \bibinfo{person}{Kang Liu}, {and}
  \bibinfo{person}{Jun Zhao}.} \bibinfo{year}{2016}\natexlab{}.
\newblock \showarticletitle{A joint embedding method for entity alignment of
  knowledge bases}. In \bibinfo{booktitle}{\emph{China Conference on Knowledge
  Graph and Semantic Computing}}. Springer, \bibinfo{pages}{3--14}.
\newblock


\bibitem[\protect\citeauthoryear{He, Zhang, Ren, and Sun}{He
  et~al\mbox{.}}{2016}]%
        {he2016deep}
\bibfield{author}{\bibinfo{person}{Kaiming He}, \bibinfo{person}{Xiangyu
  Zhang}, \bibinfo{person}{Shaoqing Ren}, {and} \bibinfo{person}{Jian Sun}.}
  \bibinfo{year}{2016}\natexlab{}.
\newblock \showarticletitle{Deep residual learning for image recognition}. In
  \bibinfo{booktitle}{\emph{Proceedings of the IEEE conference on computer
  vision and pattern recognition}}. \bibinfo{pages}{770--778}.
\newblock


\bibitem[\protect\citeauthoryear{Kim, Khanna, and Koyejo}{Kim
  et~al\mbox{.}}{2016}]%
        {kim2016examples}
\bibfield{author}{\bibinfo{person}{Been Kim}, \bibinfo{person}{Rajiv Khanna},
  {and} \bibinfo{person}{Oluwasanmi~O Koyejo}.}
  \bibinfo{year}{2016}\natexlab{}.
\newblock \showarticletitle{Examples are not enough, learn to criticize!
  criticism for interpretability}. In \bibinfo{booktitle}{\emph{Advances in
  neural information processing systems}}. \bibinfo{pages}{2280--2288}.
\newblock


\bibitem[\protect\citeauthoryear{Koh and Liang}{Koh and Liang}{2017}]%
        {koh2017understanding}
\bibfield{author}{\bibinfo{person}{Pang~Wei Koh} {and} \bibinfo{person}{Percy
  Liang}.} \bibinfo{year}{2017}\natexlab{}.
\newblock \showarticletitle{Understanding black-box predictions via influence
  functions}.
\newblock \bibinfo{journal}{\emph{arXiv preprint arXiv:1703.04730}}
  (\bibinfo{year}{2017}).
\newblock


\bibitem[\protect\citeauthoryear{Le and Mikolov}{Le and Mikolov}{2014}]%
        {le2014distributed}
\bibfield{author}{\bibinfo{person}{Quoc Le} {and} \bibinfo{person}{Tomas
  Mikolov}.} \bibinfo{year}{2014}\natexlab{}.
\newblock \showarticletitle{Distributed representations of sentences and
  documents}. In \bibinfo{booktitle}{\emph{International conference on machine
  learning}}. \bibinfo{pages}{1188--1196}.
\newblock


\bibitem[\protect\citeauthoryear{Lecue}{Lecue}{2019}]%
        {lecue2019role}
\bibfield{author}{\bibinfo{person}{Freddy Lecue}.}
  \bibinfo{year}{2019}\natexlab{}.
\newblock \showarticletitle{On the role of knowledge graphs in explainable AI}.
\newblock \bibinfo{journal}{\emph{Semantic Web}} \bibinfo{number}{Preprint}
  (\bibinfo{year}{2019}), \bibinfo{pages}{1--11}.
\newblock


\bibitem[\protect\citeauthoryear{Lee, Fang, Yeh, and Frank~Wang}{Lee
  et~al\mbox{.}}{2018}]%
        {lee2018multi}
\bibfield{author}{\bibinfo{person}{Chung-Wei Lee}, \bibinfo{person}{Wei Fang},
  \bibinfo{person}{Chih-Kuan Yeh}, {and} \bibinfo{person}{Yu-Chiang
  Frank~Wang}.} \bibinfo{year}{2018}\natexlab{}.
\newblock \showarticletitle{Multi-label zero-shot learning with structured
  knowledge graphs}. In \bibinfo{booktitle}{\emph{Proceedings of the IEEE
  conference on computer vision and pattern recognition}}.
  \bibinfo{pages}{1576--1585}.
\newblock


\bibitem[\protect\citeauthoryear{Lin, Maire, Belongie, Hays, Perona, Ramanan,
  Doll{\'a}r, and Zitnick}{Lin et~al\mbox{.}}{2014}]%
        {lin2014microsoft}
\bibfield{author}{\bibinfo{person}{Tsung-Yi Lin}, \bibinfo{person}{Michael
  Maire}, \bibinfo{person}{Serge Belongie}, \bibinfo{person}{James Hays},
  \bibinfo{person}{Pietro Perona}, \bibinfo{person}{Deva Ramanan},
  \bibinfo{person}{Piotr Doll{\'a}r}, {and} \bibinfo{person}{C~Lawrence
  Zitnick}.} \bibinfo{year}{2014}\natexlab{}.
\newblock \showarticletitle{Microsoft coco: Common objects in context}. In
  \bibinfo{booktitle}{\emph{European conference on computer vision}}. Springer,
  \bibinfo{pages}{740--755}.
\newblock


\bibitem[\protect\citeauthoryear{Lundberg and Lee}{Lundberg and Lee}{2017}]%
        {lundberg2017unified}
\bibfield{author}{\bibinfo{person}{Scott~M Lundberg} {and}
  \bibinfo{person}{Su-In Lee}.} \bibinfo{year}{2017}\natexlab{}.
\newblock \showarticletitle{A unified approach to interpreting model
  predictions}. In \bibinfo{booktitle}{\emph{Advances in neural information
  processing systems}}. \bibinfo{pages}{4765--4774}.
\newblock


\bibitem[\protect\citeauthoryear{Maaten and Hinton}{Maaten and Hinton}{2008}]%
        {maaten2008visualizing}
\bibfield{author}{\bibinfo{person}{Laurens van~der Maaten} {and}
  \bibinfo{person}{Geoffrey Hinton}.} \bibinfo{year}{2008}\natexlab{}.
\newblock \showarticletitle{Visualizing data using t-SNE}.
\newblock \bibinfo{journal}{\emph{Journal of machine learning research}}
  \bibinfo{volume}{9}, \bibinfo{number}{Nov} (\bibinfo{year}{2008}),
  \bibinfo{pages}{2579--2605}.
\newblock


\bibitem[\protect\citeauthoryear{Marino, Salakhutdinov, and Gupta}{Marino
  et~al\mbox{.}}{2016}]%
        {marino2016more}
\bibfield{author}{\bibinfo{person}{Kenneth Marino}, \bibinfo{person}{Ruslan
  Salakhutdinov}, {and} \bibinfo{person}{Abhinav Gupta}.}
  \bibinfo{year}{2016}\natexlab{}.
\newblock \showarticletitle{The more you know: Using knowledge graphs for image
  classification}.
\newblock \bibinfo{journal}{\emph{arXiv preprint arXiv:1612.04844}}
  (\bibinfo{year}{2016}).
\newblock


\bibitem[\protect\citeauthoryear{Mikolov, Sutskever, Chen, Corrado, and
  Dean}{Mikolov et~al\mbox{.}}{2013}]%
        {mikolov2013distributed}
\bibfield{author}{\bibinfo{person}{Tomas Mikolov}, \bibinfo{person}{Ilya
  Sutskever}, \bibinfo{person}{Kai Chen}, \bibinfo{person}{Greg~S Corrado},
  {and} \bibinfo{person}{Jeff Dean}.} \bibinfo{year}{2013}\natexlab{}.
\newblock \showarticletitle{Distributed representations of words and phrases
  and their compositionality}. In \bibinfo{booktitle}{\emph{Advances in neural
  information processing systems}}. \bibinfo{pages}{3111--3119}.
\newblock


\bibitem[\protect\citeauthoryear{Miller}{Miller}{1995}]%
        {miller1995wordnet}
\bibfield{author}{\bibinfo{person}{George~A Miller}.}
  \bibinfo{year}{1995}\natexlab{}.
\newblock \showarticletitle{WordNet: a lexical database for English}.
\newblock \bibinfo{journal}{\emph{Commun. ACM}} \bibinfo{volume}{38},
  \bibinfo{number}{11} (\bibinfo{year}{1995}), \bibinfo{pages}{39--41}.
\newblock


\bibitem[\protect\citeauthoryear{Mitsuhara, Fukui, Sakashita, Ogata, Hirakawa,
  Yamashita, and Fujiyoshi}{Mitsuhara et~al\mbox{.}}{2019}]%
        {mitsuhara2019embedding}
\bibfield{author}{\bibinfo{person}{Masahiro Mitsuhara},
  \bibinfo{person}{Hiroshi Fukui}, \bibinfo{person}{Yusuke Sakashita},
  \bibinfo{person}{Takanori Ogata}, \bibinfo{person}{Tsubasa Hirakawa},
  \bibinfo{person}{Takayoshi Yamashita}, {and} \bibinfo{person}{Hironobu
  Fujiyoshi}.} \bibinfo{year}{2019}\natexlab{}.
\newblock \showarticletitle{Embedding human knowledge in deep neural network
  via attention map}.
\newblock \bibinfo{journal}{\emph{arXiv preprint arXiv:1905.03540}}
  \bibinfo{volume}{5} (\bibinfo{year}{2019}).
\newblock


\bibitem[\protect\citeauthoryear{Mordvintsev, Olah, and Tyka}{Mordvintsev
  et~al\mbox{.}}{2015}]%
        {mordvintsev2015inceptionism}
\bibfield{author}{\bibinfo{person}{Alexander Mordvintsev},
  \bibinfo{person}{Christopher Olah}, {and} \bibinfo{person}{Mike Tyka}.}
  \bibinfo{year}{2015}\natexlab{}.
\newblock \showarticletitle{Inceptionism: Going deeper into neural networks}.
\newblock  (\bibinfo{year}{2015}).
\newblock


\bibitem[\protect\citeauthoryear{Pennington, Socher, and Manning}{Pennington
  et~al\mbox{.}}{2014}]%
        {pennington2014glove}
\bibfield{author}{\bibinfo{person}{Jeffrey Pennington},
  \bibinfo{person}{Richard Socher}, {and} \bibinfo{person}{Christopher~D
  Manning}.} \bibinfo{year}{2014}\natexlab{}.
\newblock \showarticletitle{Glove: Global vectors for word representation}. In
  \bibinfo{booktitle}{\emph{Proceedings of the 2014 conference on empirical
  methods in natural language processing (EMNLP)}}.
  \bibinfo{pages}{1532--1543}.
\newblock


\bibitem[\protect\citeauthoryear{Selvaraju, Cogswell, Das, Vedantam, Parikh,
  and Batra}{Selvaraju et~al\mbox{.}}{2017}]%
        {selvaraju2017grad}
\bibfield{author}{\bibinfo{person}{Ramprasaath~R Selvaraju},
  \bibinfo{person}{Michael Cogswell}, \bibinfo{person}{Abhishek Das},
  \bibinfo{person}{Ramakrishna Vedantam}, \bibinfo{person}{Devi Parikh}, {and}
  \bibinfo{person}{Dhruv Batra}.} \bibinfo{year}{2017}\natexlab{}.
\newblock \showarticletitle{Grad-cam: Visual explanations from deep networks
  via gradient-based localization}. In \bibinfo{booktitle}{\emph{Proceedings of
  the IEEE international conference on computer vision}}.
  \bibinfo{pages}{618--626}.
\newblock


\bibitem[\protect\citeauthoryear{Simonyan, Vedaldi, and Zisserman}{Simonyan
  et~al\mbox{.}}{2013}]%
        {simonyan2013deep}
\bibfield{author}{\bibinfo{person}{Karen Simonyan}, \bibinfo{person}{Andrea
  Vedaldi}, {and} \bibinfo{person}{Andrew Zisserman}.}
  \bibinfo{year}{2013}\natexlab{}.
\newblock \showarticletitle{Deep inside convolutional networks: Visualising
  image classification models and saliency maps}.
\newblock \bibinfo{journal}{\emph{arXiv preprint arXiv:1312.6034}}
  (\bibinfo{year}{2013}).
\newblock


\bibitem[\protect\citeauthoryear{Simonyan and Zisserman}{Simonyan and
  Zisserman}{2014}]%
        {simonyan2014very}
\bibfield{author}{\bibinfo{person}{Karen Simonyan} {and}
  \bibinfo{person}{Andrew Zisserman}.} \bibinfo{year}{2014}\natexlab{}.
\newblock \showarticletitle{Very deep convolutional networks for large-scale
  image recognition}.
\newblock \bibinfo{journal}{\emph{arXiv preprint arXiv:1409.1556}}
  (\bibinfo{year}{2014}).
\newblock


\bibitem[\protect\citeauthoryear{Speer, Chin, and Havasi}{Speer
  et~al\mbox{.}}{2017}]%
        {speer2017conceptnet}
\bibfield{author}{\bibinfo{person}{Robyn Speer}, \bibinfo{person}{Joshua Chin},
  {and} \bibinfo{person}{Catherine Havasi}.} \bibinfo{year}{2017}\natexlab{}.
\newblock \showarticletitle{Conceptnet 5.5: An open multilingual graph of
  general knowledge}. In \bibinfo{booktitle}{\emph{Thirty-First AAAI Conference
  on Artificial Intelligence}}.
\newblock


\bibitem[\protect\citeauthoryear{Springenberg, Dosovitskiy, Brox, and
  Riedmiller}{Springenberg et~al\mbox{.}}{2014}]%
        {springenberg2014striving}
\bibfield{author}{\bibinfo{person}{Jost~Tobias Springenberg},
  \bibinfo{person}{Alexey Dosovitskiy}, \bibinfo{person}{Thomas Brox}, {and}
  \bibinfo{person}{Martin Riedmiller}.} \bibinfo{year}{2014}\natexlab{}.
\newblock \showarticletitle{Striving for simplicity: The all convolutional
  net}.
\newblock \bibinfo{journal}{\emph{arXiv preprint arXiv:1412.6806}}
  (\bibinfo{year}{2014}).
\newblock


\bibitem[\protect\citeauthoryear{Sun, Ma, and Wang}{Sun et~al\mbox{.}}{2015}]%
        {sun2015comparative}
\bibfield{author}{\bibinfo{person}{Yufei Sun}, \bibinfo{person}{Liangli Ma},
  {and} \bibinfo{person}{Shuang Wang}.} \bibinfo{year}{2015}\natexlab{}.
\newblock \showarticletitle{A comparative evaluation of string similarity
  metrics for ontology alignment}.
\newblock \bibinfo{journal}{\emph{Journal of Information \&Computational
  Science}} \bibinfo{volume}{12}, \bibinfo{number}{3} (\bibinfo{year}{2015}),
  \bibinfo{pages}{957--964}.
\newblock


\bibitem[\protect\citeauthoryear{Sun, Hu, and Li}{Sun et~al\mbox{.}}{2017}]%
        {sun2017cross}
\bibfield{author}{\bibinfo{person}{Zequn Sun}, \bibinfo{person}{Wei Hu}, {and}
  \bibinfo{person}{Chengkai Li}.} \bibinfo{year}{2017}\natexlab{}.
\newblock \showarticletitle{Cross-lingual entity alignment via joint
  attribute-preserving embedding}. In \bibinfo{booktitle}{\emph{International
  Semantic Web Conference}}. Springer, \bibinfo{pages}{628--644}.
\newblock


\bibitem[\protect\citeauthoryear{Sun, Hu, Zhang, and Qu}{Sun
  et~al\mbox{.}}{2018}]%
        {sun2018bootstrapping}
\bibfield{author}{\bibinfo{person}{Zequn Sun}, \bibinfo{person}{Wei Hu},
  \bibinfo{person}{Qingheng Zhang}, {and} \bibinfo{person}{Yuzhong Qu}.}
  \bibinfo{year}{2018}\natexlab{}.
\newblock \showarticletitle{Bootstrapping Entity Alignment with Knowledge Graph
  Embedding.}. In \bibinfo{booktitle}{\emph{IJCAI}}.
  \bibinfo{pages}{4396--4402}.
\newblock


\bibitem[\protect\citeauthoryear{Szegedy, Liu, Jia, Sermanet, Reed, Anguelov,
  Erhan, Vanhoucke, and Rabinovich}{Szegedy et~al\mbox{.}}{2015}]%
        {szegedy2015going}
\bibfield{author}{\bibinfo{person}{Christian Szegedy}, \bibinfo{person}{Wei
  Liu}, \bibinfo{person}{Yangqing Jia}, \bibinfo{person}{Pierre Sermanet},
  \bibinfo{person}{Scott Reed}, \bibinfo{person}{Dragomir Anguelov},
  \bibinfo{person}{Dumitru Erhan}, \bibinfo{person}{Vincent Vanhoucke}, {and}
  \bibinfo{person}{Andrew Rabinovich}.} \bibinfo{year}{2015}\natexlab{}.
\newblock \showarticletitle{Going deeper with convolutions}. In
  \bibinfo{booktitle}{\emph{Proceedings of the IEEE conference on computer
  vision and pattern recognition}}. \bibinfo{pages}{1--9}.
\newblock


\bibitem[\protect\citeauthoryear{Trisedya, Qi, and Zhang}{Trisedya
  et~al\mbox{.}}{2019}]%
        {trisedya2019entity}
\bibfield{author}{\bibinfo{person}{Bayu~Distiawan Trisedya},
  \bibinfo{person}{Jianzhong Qi}, {and} \bibinfo{person}{Rui Zhang}.}
  \bibinfo{year}{2019}\natexlab{}.
\newblock \showarticletitle{Entity alignment between knowledge graphs using
  attribute embeddings}. In \bibinfo{booktitle}{\emph{Proceedings of the AAAI
  Conference on Artificial Intelligence}}, Vol.~\bibinfo{volume}{33}.
  \bibinfo{pages}{297--304}.
\newblock


\bibitem[\protect\citeauthoryear{Wang, Ye, and Gupta}{Wang
  et~al\mbox{.}}{2018b}]%
        {wang2018zero}
\bibfield{author}{\bibinfo{person}{Xiaolong Wang}, \bibinfo{person}{Yufei Ye},
  {and} \bibinfo{person}{Abhinav Gupta}.} \bibinfo{year}{2018}\natexlab{b}.
\newblock \showarticletitle{Zero-shot recognition via semantic embeddings and
  knowledge graphs}. In \bibinfo{booktitle}{\emph{Proceedings of the IEEE
  Conference on Computer Vision and Pattern Recognition}}.
  \bibinfo{pages}{6857--6866}.
\newblock


\bibitem[\protect\citeauthoryear{Wang, Lv, Lan, and Zhang}{Wang
  et~al\mbox{.}}{2018a}]%
        {wang2018cross}
\bibfield{author}{\bibinfo{person}{Zhichun Wang}, \bibinfo{person}{Qingsong
  Lv}, \bibinfo{person}{Xiaohan Lan}, {and} \bibinfo{person}{Yu Zhang}.}
  \bibinfo{year}{2018}\natexlab{a}.
\newblock \showarticletitle{Cross-lingual knowledge graph alignment via graph
  convolutional networks}. In \bibinfo{booktitle}{\emph{Proceedings of the 2018
  Conference on Empirical Methods in Natural Language Processing}}.
  \bibinfo{pages}{349--357}.
\newblock


\bibitem[\protect\citeauthoryear{Xu, Wang, Yu, Feng, Song, Wang, and Yu}{Xu
  et~al\mbox{.}}{2019}]%
        {xu2019cross}
\bibfield{author}{\bibinfo{person}{Kun Xu}, \bibinfo{person}{Liwei Wang},
  \bibinfo{person}{Mo Yu}, \bibinfo{person}{Yansong Feng}, \bibinfo{person}{Yan
  Song}, \bibinfo{person}{Zhiguo Wang}, {and} \bibinfo{person}{Dong Yu}.}
  \bibinfo{year}{2019}\natexlab{}.
\newblock \showarticletitle{Cross-lingual knowledge graph alignment via graph
  matching neural network}.
\newblock \bibinfo{journal}{\emph{arXiv preprint arXiv:1905.11605}}
  (\bibinfo{year}{2019}).
\newblock


\bibitem[\protect\citeauthoryear{Zeiler and Fergus}{Zeiler and Fergus}{2014}]%
        {zeiler2014visualizing}
\bibfield{author}{\bibinfo{person}{Matthew~D Zeiler} {and} \bibinfo{person}{Rob
  Fergus}.} \bibinfo{year}{2014}\natexlab{}.
\newblock \showarticletitle{Visualizing and understanding convolutional
  networks}. In \bibinfo{booktitle}{\emph{European conference on computer
  vision}}. Springer, \bibinfo{pages}{818--833}.
\newblock


\bibitem[\protect\citeauthoryear{Zhu, Wei, Sisman, Zheng, Faloutsos, Dong, and
  Han}{Zhu et~al\mbox{.}}{2020}]%
        {zhu2020collective}
\bibfield{author}{\bibinfo{person}{Qi Zhu}, \bibinfo{person}{Hao Wei},
  \bibinfo{person}{Bunyamin Sisman}, \bibinfo{person}{Da Zheng},
  \bibinfo{person}{Christos Faloutsos}, \bibinfo{person}{Xin~Luna Dong}, {and}
  \bibinfo{person}{Jiawei Han}.} \bibinfo{year}{2020}\natexlab{}.
\newblock \showarticletitle{Collective Multi-type Entity Alignment Between
  Knowledge Graphs}. In \bibinfo{booktitle}{\emph{Proceedings of The Web
  Conference 2020}}. \bibinfo{pages}{2241--2252}.
\newblock


\end{thebibliography}










\end{document}